\documentclass[12pt, letterpaper]{article}
\usepackage[titletoc,title]{appendix}
\usepackage{color}
\usepackage{booktabs}
\usepackage[usenames,dvipsnames,svgnames,table]{xcolor}
\definecolor{dark-red}{rgb}{0.75,0.10,0.10}
\definecolor{bluish}{rgb}{0.05,0.05,0.85}

\usepackage[margin=1in]{geometry}
\usepackage[linkcolor=black,
            colorlinks=true,
            urlcolor=black,
            pdfstartview={XYZ null null 1.00},
            pdfpagemode=UseNone,
            citecolor={black},
            pdftitle={scaling}]{hyperref}

\usepackage[resetlabels,labeled]{multibib}
\newcites{SI}{SI References}
\usepackage{natbib}

\usepackage{float}

\usepackage{geometry} 
\usepackage{graphicx}                
\usepackage{amsxtra}
\usepackage{verbatim}
\setcitestyle{round,semicolon,aysep={},yysep={;}}
\usepackage{setspace}            
\usepackage{sectsty}             
\usepackage{pdflscape}
\usepackage{fancyhdr}            
\usepackage{url}                     
\usepackage{fullpage}       
\usepackage{multirow}
\usepackage{rotating}
\usepackage[T1]{fontenc}
\usepackage[bitstream-charter]{mathdesign}

\usepackage{chngcntr}
\usepackage{booktabs}
\usepackage{longtable}

\makeatother

\usepackage{footmisc}
\makeatother

\usepackage[hang, font=small,skip=0pt, labelfont={bf}]{caption}
\usepackage{subcaption}

\usepackage{lscape}

\makeatletter
\providecommand\phantomcaption{\caption@refstepcounter\@captype}
\makeatother

\title{Scaling ML Products At Startups:\\ A Practitioner's Guide\thanks{The article benefited from comments by Pronoy Chopra, Noah Finberg, and Brian Whetter.}}

\author{Atul Dhingra\thanks{USA, \textsf{dhingra.atul92@gmail.com}} \and Gaurav Sood\thanks{USA;  \textsf{gsood07@gmail.com}.}}

\date{\today}

\begin{document}
\maketitle
\thispagestyle{empty}

\begin{abstract}

How do you scale a machine learning product at a startup? In particular, how do you serve a greater volume, velocity, and variety of queries cost-effectively? We break down costs into variable costs—the cost of serving the model and performant—and fixed costs—the cost of developing and training new models. We propose a framework for conceptualizing these costs, breaking them into finer categories, and limn ways to reduce costs. Lastly, since in our experience, the most expensive fixed cost of a machine learning system is the cost of identifying the root causes of failures and driving continuous improvement, we present a way to conceptualize the issues and share our methodology for the same.
\smallskip

\textbf{Keywords:} scaling, startups, machine learning

\end{abstract}

\section{Introduction}

As an ML startup grows, so does the volume, velocity, and variety of data it generates and uses. Handling and managing this data efficiently and effectively is critical to the success of the startup. But both premature optimization in handling this data and leaving the job till too late can mean death. To guide the thinking about which investment is needed and when, we have found it useful to divide ML startups into three stages. The first stage is the POC stage. At the POC stage, the cost is largely immaterial and the variety is moot beyond what is necessary for the POC. The second stage comports with the ability to handle modest volume, velocity, and variety, e.g., an automated contract ingestion company going from extracting data from one kind of contract to ten different contracts. In the second stage, the company needs to show progress in handling the volume, variety, and velocity of data while managing the burn rate. From the product side, the second stage can look like 10 POCs—the expected pace of delivery and the idiosyncratic nature of early customers conspire to limit serious investments in the generalization of the model. The third stage is marked by the necessity to handle massive volume, velocity, and variety of queries, e.g., an automated contract ingestion company going from 10 to 1000 kinds of contracts. In the third stage, both the speed of scaling and the cost-effectiveness of scaling are major bottlenecks to a positive return on investment. Thus, substantial investments in infrastructure are generally required to enable the company to scale quickly and provide the service cost-effectively.

\section{Cost}
The cost of a machine learning model is the sum of the cost of building (and improving) the model, maintaining the model, and serving the model. Conventionally, fixed costs, e.g., the cost of engineering the system, are excluded from consideration when scaling a digital product. Instead, it is common to exclusively focus on the contribution margin. The theory goes that once you have a positive contribution margin, profitability is a matter of acquiring and retaining enough customers. However appealing the theory sounds, it is not without its chinks. Fixed costs are important because they affect the length of the runway. Without a long enough runway, very few things take off. 

\subsection{Variable Costs}
We define variable costs as the total cost of serving the models and maintaining the models’ performance. Conventionally, companies track variable costs via metrics like variable cost per customer, variable cost per dollar earned, or percentage of Annual Recurring Revenue (ARR). Monitoring the unit cost of production provides a natural way to look for the expected trend of declining marginal costs of production. To monitor specific cost centers, it is conventional to split total cost into different cost centers, e.g., storage costs per dollar of revenue, compute costs per dollar of revenue, etc. 

Given the cost structure and cost rationalization opportunities, we split the variable costs into three broad buckets:

\begin{enumerate}

    \item \textbf{Compute costs.} Compute costs for serving large models can be substantial, especially at scale. However, any strategy aimed at reducing compute costs should price in the impact on business, which may stem from longer inference times, lower accuracy, etc. \citep{dhingra2017model}. To better enumerate potential cost-saving opportunities, we divide the compute costs into the following pieces: 
    \begin{enumerate}
        \item \textbf{The volume of queries.} All else equal, the fewer the queries, the lower the prediction costs. To highlight potential opportunities to reduce the number of queries, consider the following example. Say that you are running a startup that uses machine learning to measure the speaker’s emotions in movies. Most movies are shot at 24 frames per second. And we could set up the inference pipeline such that we predict emotions in each frame. But depending on the use case, a lower sampling rate may be enough.
    
        \item \textbf{The velocity of queries.} Serving infrastructure needs to be built with the peak rate in mind. Even with adaptive scaling, the scale-up costs can be substantial. There are at least three broad solutions. The first is forecasting. Often, if you buy computing capacity ahead of time, it is cheaper.  The second method is to constrain peak throughput. For instance, limiting the number of people who can shop at one time at a store as initial Amazon Go rollouts did (though plausibly for other reasons like controlling occlusion, theft, etc.). The last method is tiered pricing. For instance, ChatGPT offers a Pro tier that guarantees access during peak demand.  
    
        \item \textbf{Model size.} By model size, we mean the memory and the number of computations needed to make an inference. Memory Access Cost (MAC)\citep{ma2018shufflenet} and Floating Point Operations (FLOPs) are typically used to measure these costs and are often part of the model’s KPIs. The price of a larger model is generally greater latency and greater memory and compute requirements. The benefit is greater accuracy, resolution, e.g., higher resolution of the generated image, etc.
            \begin{itemize}
                \item \textbf{Memory.} The GPU memory required to load an ML model to make predictions on the target deployment platform can be affected by numerical precision, e.g., fp32, fp16, int8, the number of parameters in a model, etc. Memory requirements affect the cost for two reasons. Memory costs money. For instance, with a fixed number of FLOPs, a GPU with larger memory costs more. Second, memory can be seen as a constraint on the deployment options that are available. For instance, if the model cannot fit an on-premise device, a GPU, or a VM, you have to go with the cloud, a pricier GPU, and a pricier VM respectively.
                \item \textbf{Compute.} The total number of MAC or FLOPs needed for making a prediction. It is a function of the model capacity (the depth and the width of the network). 
        \end{itemize}
           We can reduce the model's memory and compute footprint using techniques like Knowledge Distillation \citep{hinton2015distilling}, Quantization, Graph surgery, TensorRT, etc. But it is useful to not assume the impact of these methods on latency. We have encountered cases where model distillation reduced the number of parameters by 500x but still increased the latency because of the sequence and the nature of the calls.
    \end{enumerate}

    \item \textbf{Bandwidth costs.} The cost of shuttling data for audio-visual applications can be substantial. Shuttling data generally also means greater latency and lower reliability. The standard ways of reducing bandwidth costs include compression and downsampling. For instance, the resolution of a frame that is needed for accurate prediction may be lower than the resolution at which the image or video is captured.

    \item \textbf{Operations Cost.} Operations cost pool over the cost of people and tools needed to keep the model performant. For instance, for annotation in a self-driving car company, the costs include the cost of hiring a driver, the labeling team, and the annotation tool (and its associated contract or development costs). These costs can be amortized between the cost of building new models or improving models and keeping the model performant. To reduce the operations burden, we need to optimize the data collection and annotation. To improve the speed of annotations features like predicting typing, pre-annotations, etc. can be useful. To improve the returns on the data collected, it is useful to build a rough ROI function of the type that is common in active learning. For instance, in some cases, we may want only driver miles during peak hours when the weather is bad. Other strategies include using synthetic data to drive performance gains \citep{sagers2022improving}.

\end{enumerate}

\subsubsection{Cloud Vs. On-Premise and Rent Vs. Buy}
Compute and bandwidth costs, the cost of updating the model over the air,  the cost of latency, and the cost of building and maintaining your own infrastructure, e.g., issuing security patches, monitoring and predicting hardware failures, are some of the considerations in deciding between cloud and on-premise deployment. The key virtue of the cloud is low startup costs, which is important in the era of high developer costs. It also keeps the focus on innovating in the value-generation step. Even if the cloud is the expensive option over the long term, using it initially fits with the oft-repeated “build first, optimize later” mantra.

\section{Fixed Costs}
Fixed costs cover the costs of building and improving models. The costs can be broken down into hardware costs, e.g., cars in an AV company, LIDARs, etc., and development costs, which can be further broken down into the cost of developers and the cost of model training infrastructure. The hardware costs can be amortized between variable and fixed costs based on how much of the infrastructure is used for maintaining performance and how much of it is used for building new models.

\subsection{Hardware Costs}
The sensor costs include the cost of cars, LIDARs, cameras, microphones, lighting, scanners, RFID tags, etc. It is conventional to amortize hardware expenditures based on longevity and the expected sale value when they are swapped out. Maintenance and breakage costs are generally baked into hardware costs and called out specifically under service and breakage costs. 

\subsection{Development Costs}
Development costs are a major part of the fixed costs for many companies. There are two big cost centers—the cost of the model training infrastructure and the cost of developers. 
\subsubsection{Training Infrastructure Costs}
The cost of model training infrastructure includes the cost of storage, the cost of hardware for training the models, and the cost of other engineering tools. These costs can balloon very quickly and need to be carefully modeled and projected to figure out which costs need attention.  The infrastructure can also be a bottleneck. While metrics for development velocity are less easily derived, they are probably the most important metrics. We have used self-reports through surveys to quantify the effect of better infrastructure on development speed and found that investments in data comprehension and infrastructure for data analytics for root-causing problems are generally worth it. With that, we have some specific advice about what not to do.

The five most common infrastructure-related mistakes that startups make are:
\begin{itemize}

    \item \textbf{Choosing the wrong tool for the job.} For instance, for storing user attributes for Fortnite, where not all attributes are available to all users, using RDS than DDB. when s3 is the right call. The root cause could be a bias for building with what you are familiar with and unawareness of reference architectures. The solution is taking the time to educate yourself about the reference architecture and the use case. 
    
    \item \textbf{Reinventing tools.} Given the developer costs, inventing developer tools for ML startups should be an option of the last resort till well into the third stage or unless the math really tallies in favor of such an investment. The thumb rule is that no developer problem is novel. If your company is struggling with a problem, it is likely that other companies are too. And major cloud service providers or another company generally already has some answer for that problem. Add to it the point that rarely is the ROI on the perfect solution substantially greater than a good enough solution. For instance, SageMaker can be an effective way to deploy models till the company is well into the third stage. Lastly, when picking cloud tools, another of the maxims is: when in doubt, go serverless.
    
    \item \textbf{Not carefully modeling how costs will increase with scale.} The growth in costs in infrastructure is often non-linear. Before the marginal costs come down, they often increase. If you do not carefully model the costs, the rise in costs will likely come as a shock, leaving little time for crafting an effective mitigation strategy.  
    
    \item \textbf{Ignorance of cost optimization opportunities.} Simple principles of infrastructure savings—volume discounting, e.g., using forecasting to buy compute in bulk, auto-scaling, using resources during times when they are cheaper, e.g., training costs can be brought down by 80—90\% by using spot instances but require the use of checkpointing, exist across the board. 
    
    \item \textbf{False Optimization.} Companies sometimes just move the pain than save money. For instance, some companies do not enable logs because of concerns about storage expenses. (These concerns can be overcome by setting up ways to archive and eventually delete the logs.) And the consequence is that it is much harder (and takes much longer) to diagnose errors.

\end{itemize}

\subsubsection{Developer costs}

Developer costs can be split into three kinds of costs:
\begin{itemize}
    \item \textbf{The cost of engineers.} All else equal, the number of engineers required to serve the customers at a particular quality grows sharply with the diversity, volume, and velocity of the data. The only way to reduce the slope of the function is by increasing the maturity of the infrastructure and paying down the technical debt. Till the second stage, reducing the slope may not be optimal but in the third stage, it becomes sine-qua-non.

The cost of engineering churn. The cost is a sum of two costs:
\begin{itemize}
    \item  \textbf{The cost of losing tribal knowledge.} At startups, many critical pieces of how the system works reside within people and good documentation is generally unavailable. Losing an engineer means losing a lot of the context which makes development much slower and also means that engineers have to relearn the same lessons which can be expensive (and time-consuming). Given that startups cannot often afford redundancy—multiple engineers responsible for the same piece of code—in the extremum, an engineer leaving can have dramatic consequences for the maintenance of a particular feature. 

    \item \textbf{Hiring costs.} Hiring is expensive and time-consuming. It costs the remaining engineers interview time. It means generally that the company has access to one less engineer so the delivery speed is slower than what is optimal. It also includes the cost of ramping up another engineer.  Add to all of this the fact that all else equal, losing an engineer increases the odds of another engineer leaving. 
\end{itemize}

There are many reasons for regrettable engineering churn. Aside from compensation, and the conditions in the wider job market, some engineers leave because of poor quality of life. Some common aspects of startup life—the amount of work, frequent reorganizations, and frequent movement across workstreams—make working at a startup more unpleasant. Beyond that, as the company moves to the third stage, engineers can be increasingly stuck fixing bugs if the company has not paid down the technical debt. And it is a canon among engineers that fixing bugs is not as satisfying as building new features. Engineers also generally think it is particularly unappealing to fix other people’s bugs and to fix ‘simple’ bugs that come with a large overhead. Because of these reasons, for many engineers, the more time spent fixing bugs, the less pleasant the job. The time devoted to fixing bugs is some function of the quality of the code and infrastructure, and the quality of tools used to root cause errors. For all of these reasons, paying down technical debt should be moved forward. We discuss how to think about technical debt in the next section.

\item \textbf{The cost of lower product quality.} The cost of a lower-quality product is customer churn, poor reputation, and lower sales. One measure of product quality is the number of customer complaints (multiplied by their impact—often crudely coded into simple categories, e.g., high, medium, and low—and the time for which they remain unsolved) that stem from bugs. This metric has the virtue of aligning quality in ways that are important to the customer. It is also conventional to add uptime as a KPI. 

Nearly all software ships with bugs and often the bugs are known—the bugs are there because of agreed-upon trade-offs, which often go undocumented. So when using data on bugs to make a decision about whether or not to invest in operational excellence, it is useful to focus on unanticipated bugs.

In our experience, the number of unanticipated bugs is weakly related to conventional code quality metrics, e.g., the extent of logging, the number and coverage of tests, the quality of documentation, etc. The largest predictor is the quality of engineers and oversight.   

\item \textbf{The cost of lower development velocity.} The consequences of a lower cadence of feature delivery and longer times to deploy to a new site or customer are stark—poorer sales and greater cash burn (and the associated risk of running out of money). The conventional metrics for tracking this problem are
    \begin{itemize}
        \item \textbf{The frequency of releases.} (Or the rate of feature delivery, pro-rated for complexity). Many big engineering problems eventually cause the release cadence to become slower. The rate of production needs to be broken down into its own input metrics. For instance, are machine learning engineers spending most of their time wrangling data? Is a lack of data versioning what is causing the release process to fail?
    
    \item \textbf{The time to release to a new customer or site.} Over time, we expect this metric to sharply come down. This is a metric that needs to be broken down and understood for scaling. 
    \end{itemize}
\end{itemize}

\subsubsection{Technical Debt}
Startups are financed by technical debt. It is generally imprudent to aim for zero technical debt. The level at which you engineer solutions should reflect the needs and constraints of the time. For instance, if you are testing your ETAs for a route planning algorithm in a second-stage company, you may write a script against Google map routing. Eventually, as the needs evolve, you may want to write a class or add the ability to query other mapping services. Similarly, at the start of some ML projects where the data generation process is expected to not change,  it may not be useful to monitor performance against fresh samples as designing and maintaining such systems is expensive. Instead, like other engineering solutions, e.g., regular-expression-based systems, it may be reasonable to rely on customer complaints as a way to flag problems. On the flip side, it pays to take on the right kind of debt and pay down the high-interest debt. So how do we think about which technical debt should be paid down first? The core principles are:
\begin{itemize}
    \item \textbf{Pay down the most expensive debt first.} The cost can be customer value or the cost of maintenance.
    \item \textbf{Solving upstream technical debt is generally more important than downstream technical debt.} In ML, paying down data quality issues is better than building solutions to deal with data quality in the model. For instance, fixing sensors or their integration to get better data often provides a higher return on investment.
\end{itemize}

Not all technical debt is a result of engineering to the needs or constraints of the time. A lot of it is plain bad engineering practices. Debt avoided is money saved. Following the meta engineering principles for writing maintainable software, software that is easily operable, e.g., easy to troubleshoot and run, simple, e.g., which can be achieved through abstraction, etc., can save a lot of money. 

\section{Maintaining Performance}
Maintaining performance over time is a three-step process: 1. defining and monitoring performance, 2. root-causing failures, and 3. finding and deploying the fixes. The last part is too problem specific so we limit our discussion to the first two points.

\subsection{Defining and Tracking Performance}
Maintaining performance starts with defining performance objectives that capture all relevant costs to the business. For instance, the performance of a credit card fraud detection algorithm can be defined using two metrics the sum of the cost of fraud (a consequence of false negatives—transactions that were fraudulent but not flagged) and the cost of bad customer experience (a consequence of false positives—transactions where no fraud was committed but that were flagged).  The costs need to be comprehensive. For instance, the cost of a bad customer experience needs to take into account direct and indirect costs, e.g., the cost of resolving the issue, the impact on customer retention, and the impact on business reputation. The other Key Performance Indicators (KPIs) may include the latency of the model and the cost of serving a prediction.

When monitoring metrics, it is easy to be misled by small spikes. One way to avoid fixating on false positives is to smooth the time series using an exponentially weighted moving average or Kalman filter. Smoothing, however, comes at the cost of lower sensitivity. One way to increase the sensitivity is to increase the sample size. Larger samples are also essential if you want a sensitive measure for various slices of the data. Another reason for spikes is (expected) variation from seasonality and such. For instance, for an ice cream manufacturer, it may be reasonable to adjust sales for seasonal variation. To the extent that variables like seasonality reflect factors outside the control (as in the example above rather than issues like poor posture recognition of people in winter clothes), it is useful to regress those out. 

\subsection{Dashboards}

To triage performance, we need to regress out things that are known so that we can focus on unknown unknowns. For that reason, dashboards should reflect all known sources of variation in the performance metric. Say that we know that we perform worse on occluded images. We can account for the issue by adjusting for the proportion of occluded images or by showing performance on occluded and unoccluded images and having a panel that tracks the proportion of occluded images over time. Another helpful way of laying out the dashboard is to lay it out as a funnel that roughly maps to the system diagram, with each output metric split into input metrics. 

\subsection{Identifying the Root Causes of Problems}
There are three parts to problem-solving: 1. Coming up with a potential set of explanations, 2. Putting priors on the explanations, and 3. ruling out rival explanations. We can generate explanations by analyzing the funnel, e.g., where the dropoff is happening, sampling failures, and looking at correlations,e.g., performance is worse in a certain subgroup. To come up with priors, it is common to use prior experience or principles like proximal explanations are more likely. One way to prioritize the search is to go in order of probability though often it is useful to think about some of the probable causes and see if you can exclude some. One way to isolate causes is to look for which explanation explains all the data. The reason this strategy works is that there is often only a single point of failure. For other strategies, see \cite{shroff}.

\subsection{Model Troubleshooting}
Traditional software engineering practice embraces strong abstraction boundaries using encapsulation and modular design to create maintainable code in which it is easy to make isolated changes and improvements. But ``machine learning models are machines for creating entanglement and making the isolation of improvements effectively impossible.'' IML systems follow the ``CACE principle: Changing Anything Changes Everything'' \citep{sculley2015hidden}. This is partly why explainable AI that promises to explain why the model predicted X vs. Y is considered a promising tool for understanding problems \citep{duckworth}. But we have a long way to go with explainable AI and in our ability to come up with good causal explanations. For that reason, it helps to have a conceptual typology of why models fail.

\subsubsection{Typology of Conceptual Issues That Cause Models to Fail}

Models fail for six broad conceptual reasons:

\begin{enumerate}

    \item \textbf{Bad data.} By bad data, we mean cases where tooling or coding errors corrupt the data. For instance, an aggregation pipeline not accounting for missing data on one of the feature vectors and produces Nans.

    \item \textbf{Wrong y.} Cases where the data generation process for the dependent variable changed such that the assumptions used to build the model no longer hold. Say, for instance, you bootstrap a recommendation model using click data from randomly picked news articles. Say that the model is used to generate the set of articles that the user sees and what the users click on then is used to further refine the model. In this case, the data generation process is continuously changing as the last period of data affects what articles people see and click on. One of the common risks here is showing a very narrow category of articles to the user. This can cause a large deterioration in our ability to learn and optimize for user preferences, which may change over time.

    \item \textbf{Missing features.} Features limit what can be learned. If say you are building a housing price predictor and do not have data on road noise that can be heard from the house, and say that is an important predictor of housing prices and is weakly correlated with other features, then there is little chance of reducing the error stemming from the omitted variable.

    \item \textbf{Limited representation.} Even when the variable is present, the representation of the feature (data) limits what relationships can be learned. For instance, in sentiment classification, the order of words matters. If you use a bag of words model, there is no way to learn about sequences.
    
    \item \textbf{Extrapolation error.} The kind of data that the model is predicting on is different from what it was trained on. And that means the assumptions behind the model no longer hold.

   \item  \textbf{Interpolation error.} Interpolation error stems from two sources—(local) underfitting and (local) overfitting. There are multiple sources of interpolation errors. First, cross-validation, the primary tool we use to prevent overfitting is too crude a tool to prevent local overfitting. Generally, the model losses are too insensitive to local overfitting and underfitting across narrow slices of data. The second source of issues is model capacity. If you have a linear model and the true function is quadratic, it will lead to an interpolation error. One of the solutions to both local overfitting and underfitting is to get more data. To fix underfitting to the data, it is helpful to increase model capacity. 
\end{enumerate}

\subsubsection{Detecting Model Problems}
Like other logic built on the assumption that data distribution is stationary, e.g., Regex, machine learning applications can fail silently. Any change in the data-generating process needn't affect the business metrics or indeed even the model's topline metrics but it is useful to be aware of the shifts. Here are some ways to measure the consequences of changes in the data-generating process: 
\begin{enumerate}

    \item \textbf{Error on annotated data.} In supervised learning applications, one way to diagnose if there is a problem is to continuously annotate new data. To trigger an investigation, one heuristic is that performance on the latest batch is worse than previous batches. Usually, we only annotate a small sample each time period, and sampling variability alone could cause the metrics to fluctuate. To address that, we can smooth the estimates and adjust them for known-knowns. For instance, if you have a model that predicts the ETA of the truck, you may want to adjust for scenarios where the prediction is known to be worse, e.g., the proportion of truck loads that are scheduled to pick up a load during peak hours.
    
    \item \textbf{Distribution of model prediction confidence intervals.} Detect drift indirectly by nonparametrically testing the width of the confidence intervals.
    
    \item \textbf{Out of range.} Monitor the entry of new classes (or in the case of continuous variables, whether the values are beyond the previously observed range) in the dependent and the independent variables. 
    
    \item \textbf{Concept drift.} Is the composition of a class changing? The ways to detect it include monitoring the covariance, correlation, clusters, etc. in a class.
    
    \item \textbf{Data drift.} In our experience, monitoring generic aspects of the covariates has not been particularly fruitful in detecting or root-causing errors. But it is a good data hygiene practice. Methods like tracking reconstruction error have proven more effective \citep{zavrtanik2021reconstruction}.

\end{enumerate}

\subsection{Tools}

Different tools are needed for different scales. For instance, companies rarely invest in internal tools at the first stage. In the second stage, companies focus on tools that help you dive deep into the product vertical you are trying to capture, e.g., an AV company, which has completed a successful POC on a single state highway will likely need to invest in data exploration tools to solve L4 automation for all US highways. However, once companies surpass the second stage, they need a way to store, represent, and visualize the data in a more indexable manner. Typically, you need a feature store to not just capture the variance within data, but also to be able to replicate model degradation on a large, growing set of input data.

\section{Conclusion}
Scaling startups is hard. Challenges unique to machine learning make scaling machine learning startups harder. In this paper, we provide a framework for thinking about costs and highlight some cost saving opportunities. We also outline ways to think about one of the most expensive portions of machine learning---root causing problems. 

\newpage
\bibliographystyle{apsr}
\bibliography{scaling}

@article{hinton2015distilling,
  title={Distilling the knowledge in a neural network},
  author={Hinton, Geoffrey and Vinyals, Oriol and Dean, Jeff},
  journal={arXiv preprint arXiv:1503.02531},
  year={2015}
}

@article{zavrtanik2021reconstruction,
  title={Reconstruction by inpainting for visual anomaly detection},
  author={Zavrtanik, Vitjan and Kristan, Matej and Sko{\v{c}}aj, Danijel},
  journal={Pattern Recognition},
  volume={112},
  pages={107706},
  year={2021},
  publisher={Elsevier}
}

@article{sculley2015hidden,
  title={Hidden technical debt in machine learning systems},
  author={Sculley, David and Holt, Gary and Golovin, Daniel and Davydov, Eugene and Phillips, Todd and Ebner, Dietmar and Chaudhary, Vinay and Young, Michael and Crespo, Jean-Francois and Dennison, Dan},
  journal={Advances in neural information processing systems},
  volume={28},
  year={2015}
}

@article{dhingra2017model,
  title={Model complexity-accuracy trade-off for a convolutional neural network},
  author={Dhingra, Atul},
  journal={arXiv preprint arXiv:1705.03338},
  year={2017}
}

@article{sagers2022improving,
  title={Improving dermatology classifiers across populations using images generated by large diffusion models},
  author={Sagers, Luke W and Diao, James A and Groh, Matthew and Rajpurkar, Pranav and Adamson, Adewole S and Manrai, Arjun K},
  journal={arXiv preprint arXiv:2211.13352},
  year={2022}
}

@article{shroff,
  title={Problem Solving},
  author={Shroff, Sid and Sood, Gaurav},
  year={2023}
}

@inproceedings{ma2018shufflenet,
  title={Shufflenet v2: Practical guidelines for efficient cnn architecture design},
  author={Ma, Ningning and Zhang, Xiangyu and Zheng, Hai-Tao and Sun, Jian},
  booktitle={Proceedings of the European conference on computer vision (ECCV)},
  pages={116--131},
  year={2018}
}

@article{duckworth,
	author = {Duckworth, Christopher and Chmiel, Francis P. and Burns, Dan K. and Zlatev, Zlatko D. and White, Neil M. and Daniels, Thomas W. V. and Kiuber, Michael and Boniface, Michael J.},
	date = {2021/11/26},
	doi = {10.1038/s41598-021-02481-y},
	id = {Duckworth2021},
	isbn = {2045-2322},
	journal = {Scientific Reports},
	number = {1},
	pages = {23017},
	title = {Using explainable machine learning to characterise data drift and detect emergent health risks for emergency department admissions during COVID-19},
	url = {https://doi.org/10.1038/s41598-021-02481-y},
	volume = {11},
	year = {2021}}

\end{document}